# Speckle Noise Reduction in Medical Ultrasound Images

Faouzi Benzarti,  Hamid Amiri

Signal, Image and Pattern Recognition Laboratory (TSIRF) ENIT
Engineering School of Tunis (ENIT)
Benzartif@yahoo.fr

**Abstract**
Ultrasound imaging is an incontestable vital tool for diagnosis, it provides in non-invasive manner the internal structure of the body to detect eventually diseases or abnormalities tissues. Unfortunately, the presence of speckle noise in these images affects edges and fine details which limit the contrast resolution and make diagnostic more difficult. In this paper, we propose a denoising approach which combines logarithmic transformation and a non linear diffusion tensor. Since speckle noise is multiplicative and nonwhite process, the logarithmic transformation is a reasonable choice to convert signal-dependent or pure multiplicative noise to an additive one. The key idea from using diffusion tensor is to adapt the flow diffusion towards the local orientation by applying anisotropic diffusion along the coherent structure direction of interesting features in the image. To illustrate the effective performance of our algorithm, we present some experimental results on synthetically and real echographic images.

## 1. Introduction

In the last few decades, several non-invasive new imaging techniques have been discovered such as CT scan, SPECT, ultrasound, digital radiography, magnetic resonance imaging (MRI), spectroscopy and others. These techniques have revolutionized diagnostic radiology, providing the clinician with new information about the interior of the human body that has never been available before. Among these imaging techniques, we are interested on the ultrasound imaging which is a popular non invasive and low cost technique to observe the dynamical behavior of organs. This technique uses ultrasonic waves which are produced from the transducer and travel through body tissues. The return sound wave vibrates the transducer which turns into electrical pulses that travel to the ultrasonic scanner where they are processed and transformed into a digital image [4]. The resolution of the image will be better by using higher frequencies but this limits the depth of the penetration. However, the presence of noise is imminent due to the loss of proper contact or air gap between the transducer probe and body [28]. Speckle is a particular type of noise which degrades the fine details and edges definition and limits the contrast resolution by making it difficult to detect small and low contrast lesions in body. The challenge is to design methods which can selectively reducing noise without altering edges and losing significant features. Several methods have been proposed to remove speckle noise including temporal averaging homomorphic Wiener filtering [18]. temporal averaging [27], median filtering[25], adaptive speckle reduction [26] and wavelet Thresholding [24]. Adaptive filters [8][9][13] have the advantage that they take into account the local statistical properties of the image in which speckle noise can be well reduced but small details tend to be lost. In the past few years, the use of non linear PDEs methods involving anisotropic diffusion has significantly grown and becomes an important tool in contemporary image processing. The key idea behind the anisotropic diffusion is to incorporate an adaptative smoothness constraint in the denoising process. That is, the smooth is encouraged in a homogeneous region and discourage across boundaries, in order to preserve the discontinuities of the image. One of the most successful tools for image denoising is the Total Variation (TV) model proposed by Rudin and al [10] [11][12] and the anisotropic smoothing model proposed by Perona and Malik [1], which has since been expanded and improved upon [7][2]. Over the years, other very interesting denoising methods have been emerged such as: Bilateral filter and its derivatives [17]. In our work, we address ultrasound images denoising by using nonlinear diffusion tensor derived from the so-called structure tensors which have proven their effectiveness in several areas such as: texture segmentation, motion analysis and corner detection [7][16][14], The structure tensor provides a more powerful description of local patterns images better than a simple gradient. Based on its eigenvalues and the corresponding eigenvectors, the tensor summarizes the predominant directions of the gradient in a specified neighborhood of a point, and the degree to which those directions are coherent.

This paper is organized as follows. Section 2, introduces the non linear diffusion PDEs and discuss the various options that have been implemented for the anisotropic diffusion. Section 3, depicts the non linear structure tensor formalism and its mathematical concept. Section 4, focuses on the proposed denoising approach. Numerical experiments and comparison with some existing denoising filters results are given in Section 5.

## 2. Non linear diffusion PDEs

In this section, we review some basic mathematical concepts of the nonlinear diffusion PDEs and their application in the denoising process. Let $u(x, y, t): \Omega \to R$ be the grayscaled intensity image with a diffusion time t, for the image domain $\Omega \in R^2$. The nonlinear PDE equation is given by [1]:

$$\partial_t u = div(g(|\nabla u|)\nabla u) \quad on\ \Omega \times (0, \infty) \quad (1)$$
$$u(x, y, 0) = u_0 \quad on\ \Omega \quad \text{(e.g. Initial Condition)}$$
$$\partial u_n = 0\ on\ \partial\Omega \times (0, \infty) \quad \text{(e.g. reflecting boundary)}$$

Where $\partial_t u$: denotes the first derivative regarding the diffusion time t; $|\nabla u|$: denotes the gradient modulus and g(.) is a non-increasing function, known as the diffusivity function which allow isotropic diffusion in flat regions and no diffusion near edges. By developing the divergence term of (1), we obtain [3]:

$$\partial_t u = g''(|\nabla u|)u_{\eta\eta} + \frac{g'(|\nabla u|)}{|\nabla u|}u_{\xi\xi} \quad (2)$$
$$= c_\eta u_{\eta\eta} + c_\xi u_{\xi\xi}$$

where: $u_{\eta\eta} = \eta^\perp H\ \eta$ and $u_{\xi\xi} = \xi^\perp H\ \xi$ are respectively the second spatial derivatives of u in the directions of the gradient $\eta = \nabla u/|\nabla u|$, and its orthogonal $\xi = \eta^\perp$; H denotes the Hessian of u. According to these definitions, we have diffusion along $\eta$ (normal to the edge) weighted with $c_\eta = g''(|\nabla u|)$ and a diffusion along $\xi$ (tangential to the edge) weighted with $c_\xi = g'(|\nabla u|)/|\nabla u|$. To understand the principle of the anisotropic diffusion, let represent a contour **C** (figure 1) separating two homogeneous regions of the image, the isophote lines (e.g. level curves of equal gray-levels) correspond to u(x,y) = c. In this case, the vector $\eta$ is normal to the contour **C**, the set $(\xi, \eta)$ is then a moving orthonormal basis whose configuration depends on the current coordinate point (x, y). In the neighborhood of a contour **C**, the image presents a strong gradient. To better preserve these discontinuities, it is preferable to diffuse only in the direction parallel to **C** (i.e. in the $\xi$ -direction). In this case, we have to inhibit the coefficient of $u_{\eta\eta}$ (e.g. $c_\eta = 0$), and to suppose that the coefficient of $u_{\xi\xi}$ does not vanish.

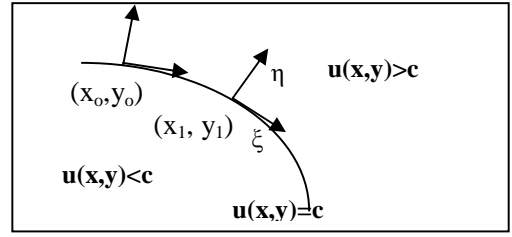

**Fig.1** Image contour and its moving orthonormal basis $(\xi, \eta)$

So it appears the following conditions for the choice of the g( ) functions to be an edge preserving :
- $g''(0) \geq 0\ and\ g'(0) \geq 0$ : avoids inverse diffusion
- $\lim_{|\nabla u| \to 0} c_\eta = \lim_{|\nabla u| \to 0} c_\xi = cte > 0$ : allows isotropic diffusion for low gradient.
- $\lim_{|\nabla u| \to \infty} c_\eta = \lim_{|\nabla u| \to \infty} c_\xi = 0\ and\ \lim_{|\nabla u| \to \infty} \frac{c_\eta}{c_\xi} = 0$:

allows anisotropic diffusion to preserve discontinuities for the high gradient. Among functions satisfying these conditions, the exponential function [1]:

$$g(|\nabla u|) = e^{-\frac{|\nabla u|^2}{k}} \quad (3)$$

This function introduces a parameter *k* which acts as a threshold which determines whether to preserve edges or not. Areas in which the gradient magnitude is lower than *k* will be blurred more strongly than areas with a higher gradient magnitude. This tends to smooth uniform regions, while preserving the edges between different regions. However, the parameter *k* can lead to a backward diffusion, when its value is smaller than $|\nabla u|$, and thus altering edges. It is evident that the optimal value for the parameter *k* has to depend on the problem. However, the anisotropic diffusion process in (1) does not perform well for large noisy images. The problem is mainly due to the dependence of the variable diffusion coefficient $g(|\nabla u|)$ on the magnitude of the image gradient. If the initial image *u* is very noisy, then large oscillations in $(|\nabla u|)$ will result in a large number of false edges. To avoid this instability, a smoothed version of the image gradient is usually used; the gradient of *u* is replaced by:

$$\nabla u_\delta = \nabla(G_\delta * u) \quad (4)$$

Where $G_\delta$ is a Gaussian with standard deviation $\delta$.

However, the non linear diffusion PDE approach does not give reliable information in the presence of local orientation structures (e.g. fingerprints).

## 3. Structure Tensor

In order to identify features such as corners or to measure the local coherence of structures, it is necessary using methods which take into account the orientation of the gradient and the flow towards the orientation of interesting features. This can be easily achieved by using the structure tensor, also referred to the second moment matrix. For a multivalued image, the structure tensor has the following form [6]:

$$S_\sigma = \left(\sum_{i=1}^n \nabla u_{i\sigma} \nabla u_{i\sigma}^T\right) = \begin{bmatrix} \sum_{i=1}^n u_{ix\sigma}^2 & \sum_{i=1}^n u_{ix\sigma} u_{iy\sigma} \\ \sum_{i=1}^n u_{ix\sigma} u_{iy\sigma} & \sum_{i=1}^n u_{iy\sigma}^2 \end{bmatrix} \quad (5)$$

With $\nabla u_{i\sigma} = K_\sigma * \nabla u_i = K_\sigma * (u_{ix}, u_{iy})$ : the smoothed version of the gradient which is obtained by a convolution with a Gaussian kernel $K_\sigma$.

The structure scale $\sigma$ determines the size of the resulting flow-like patterns. Increasing $\sigma$ gives an increased distance between the resulting flow lines. These new gradient features allow a more precise description of the local gradient characteristics. However, it is more convenient to use a smoothed version of $S_\sigma$, that is:

$$J_\rho = K_\rho * S_\sigma = \begin{bmatrix} j_{11} & j_{12} \\ j_{21} & j_{22} \end{bmatrix} \quad (6)$$

Where $K_\rho$ : a Gaussian kernel with standard deviation $\rho$.

The integration scale $\rho$ averages orientation information. Therefore, it helps to stabilize the directional behavior of the filter. In particular, it is possible to close interrupted lines if $\rho$ is equal or larger than the gap size. In order to enhance coherent structures, the integration scale $\rho$ should be larger than the structure scale $\sigma$ ($ex: \rho = 3\sigma$). In summary, the convolution with the Gaussian kernels $K_\sigma, K_\rho$, make the structure tensor measure more coherent. To go further into the formalism, the structure tensor $J_\rho$ can be written over its eigenvalues $(\lambda_+, \lambda_-)$ and eigenvectors $(\theta_+, \theta_-)$, that is [2]:

$$J_\rho = (\theta_+ \ \theta_-) \begin{bmatrix} \lambda_+ & 0 \\ 0 & \lambda_- \end{bmatrix} \begin{bmatrix} \theta_+ \\ \theta_- \end{bmatrix} = \lambda_+ \theta_+ \theta_+^T + \lambda_- \theta_- \theta_-^T \quad (7)$$

The eigenvectors of $J_\rho$ give the preferred local orientations, and the corresponding eigenvalues denote the local contrast along these directions. The eigenvalues of $J_\rho$ are given by:

$$\lambda_+ = \frac{1}{2}(j_{11} + j_{22} + \sqrt{(j_{11} - j_{22})^2 + 4j_{12}^2})$$

$$\lambda_- = \frac{1}{2}(j_{11} + j_{22} - \sqrt{(j_{11} - j_{22})^2 + 4j_{12}^2}) \quad (8)$$

And the eigenvectors $(\theta_+, \theta_-)$ satisfy :

$$\theta_+ \| \begin{pmatrix} \frac{2j_{12}}{\sqrt{(j_{22}-j_{11}+\sqrt{(j_{11}-j_{22})^2+4j_{12}^2})^2+4j_{12}^2}} \\ \frac{(j_{22}-j_{11}+\sqrt{(j_{11}-j_{22})^2+4j_{12}^2})}{\sqrt{(j_{22}-j_{11}+\sqrt{(j_{11}-j_{22})^2+4j_{12}^2})^2+4j_{12}^2}} \end{pmatrix} \quad (9)$$

and $\theta_- \perp \theta_+$.

The eigenvector $\theta_+$ which is associated to the larger eigenvalue $\lambda_+$ defines the direction of largest spatial change (i.e. the "gradient" direction). There are several ways to express the norm of the vector gradient to detect edges and corners; the most used is $N = \sqrt{\lambda_+ + \lambda_-}$. The eigenvalues $(\lambda_+, \lambda_-)$ are indeed well adapted to discriminate different geometric cases:

- If $\lambda_+ \cong \lambda_- \cong 0$, the region doesn't contain any edges or corners. For this configuration, the variation norm $N$ should be low.
- If $\lambda_+ \gg \lambda_-$ there are a lot of vector variations. The current point may be located on a vector edge. For this configuration, the variation norm $N$ should be high.
- If $\lambda_+ \cong \lambda_- \gg 0$, There is a saddle point of the vector surface, which can possibly be a vector corner in the image In this case $N$ should be even higher than the case above.

Hence, it clearly appears that the structure tensor is well suited for estimating local orientation and closing interrupted lines in coherent flow like textures.

## 4. Proposed Approach

The proposed method consists of four steps as summarized in figure 2. The first step is necessary when the degraded image is very low contrast. This can be accomplished by using the histogram equalization which allows areas of lower local contrast to gain a higher contrast.

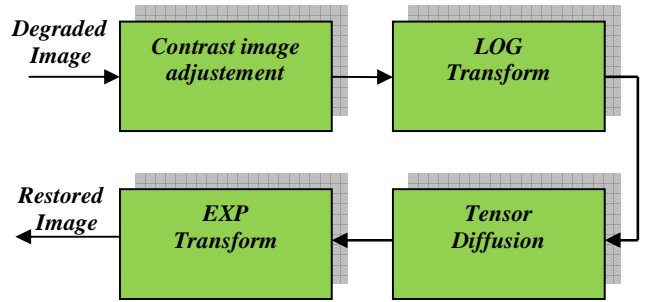

**Fig.2** Block diagram of the proposed method

Since speckle noise is multiplicative, the logarithmic transformation is a reasonable choice to convert it to an additive one. By neglecting additive gaussian noise, we have:

$$f(x,y) = \eta_s f_0(x,y) \quad (10)$$

Where : $f(x,y)$: noisy image, $f_0(x,y)$: noiseless original image, $\eta_s$: multiplicative noise.
The Log transform leads then to the following equation:

$$u(x,y) = \log(f(x,y)) = \log(f_0(x,y)) + \log(\eta_s) \quad (11)$$

However, the proposed approach is mainly based on the non linear diffusion tensor derived from the structure tensor of equation (6). The idea is to allow diffusion along the orientation of greatest coherence. The concept of the non linear diffusion tensor is obtained by replacing the diffusivity function $g()$ in (1) with a structure tensor, to create a truly anisotropic scheme, that is [5]:

$$\partial_t u = \mathrm{div}(D(J_\rho)\nabla u) \quad (12)$$

where $D(.)$ is the diffusion tensor which is positive definite symmetric 2x2 matrix. This tensor posses the same eigenvectors $\theta_-$, $\theta_+$ as the structure tensor $J_\rho$ and uses $\lambda_1$ and $\lambda_2$ to control the diffusion speeds in these two directions, that is :

$$D(J_\rho) = \begin{bmatrix} a & b \\ b & c \end{bmatrix} = (\theta_+ \ \theta_-)\begin{bmatrix} \lambda_1 & 0 \\ 0 & \lambda_2 \end{bmatrix}\begin{bmatrix} \theta_+ \\ \theta_- \end{bmatrix} = \lambda_1 \theta_+ \theta_+^T + \lambda_2 \theta_- \theta_-^T \quad (13)$$

The diffusion tensor D takes the form of an ellipsoid as is represented in figure 3. We note that for the scalar image from (1), the diffusion tensor is reduced to $D = g(|\nabla u|)I_d$ ; where $I_d$ : Identity matrix.

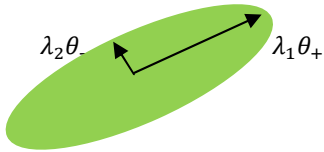

**Figure 3** Diffusion tensor 2D representation

However, several choices for the diffusion weight functions $\lambda_1$ and $\lambda_2$ are possible, depending on the considered application. For our denoising problem, we propose using:

$$\begin{cases} \lambda_1 = \dfrac{C}{\sqrt{1+N}} \\ \lambda_2 = c.\sqrt{\dfrac{1+\lambda_+}{1+\lambda_-}} \end{cases} \quad (14)$$

Where $N = \sqrt{\lambda_+ + \lambda_-}$, $C$ is a real positive parameters.
- In flat regions, we should have $\lambda_+ = \lambda_- = 0$, and then $\lambda_1 = \lambda_2 = C$ ; $D = C\,I_d$ where $I_d$ is the identity matrix. The tensor D is defined to be isotropic in these regions and takes the form of a circle of radius C.
- Along image contours, we have $\lambda_+ \gg \lambda_- \gg 0$, and then $\lambda_2 > \lambda_1 > 0$. The diffusion tensor D is then anisotropic, mainly directed by the smoothed direction $\theta_-$ of the image isophotes.

As a result, broken boundaries of a single structure could be reconnected by allowing diffusion along the orientation of greatest coherence. Furthermore, equation (12) can be solved numerically using finite differences approximation [7]. The time derivative $\partial_t u$ at $(i,j,t_n)$ is approximated by the forward difference $\partial_t u = (u^{n+1} - u^n)/\Delta t$, which leads to the iterative scheme:

$$u^{n+1} = u^n + \Delta t\, \mathrm{div}(D(J_\rho)\nabla u^n) \quad (15)$$

And the divergence term is approximated using a symmetrical central difference that is [14]:

$$\mathrm{div}(D\nabla u) = \partial_x(a\partial_x u + b\partial_y u) + \partial_y(b\partial_x u + c\partial_y u) \quad (16)$$
$$= \partial_x(a\partial_x u) + \partial_x(b\partial_y u) + \partial_y(b\partial_x u) + \partial_y(c\partial_y u)$$

with $\partial_x F = \frac{1}{2}(F_{i+1,j} - F_{i-1,j})$, $\partial_y F = \frac{1}{2}(F_{i,j+1} - F_{i,j-1})$, $F$ : represents any terms

Finally, we have to apply the exponential transform to obtain the restored image in the linear coordinates space (x,y). In the following section, we will give some experimental results.

## 5. Experimental results

We first test the performance of the proposed algorithm with a synthetically image having many oriented curves sized 256x256 pixels figure 4-(a). A speckle noise with variance $\sigma^2 = 0.02$ is added to the original image to obtain a noisy version Figure 4(b). The model's parameters are fixed to: Nb iter=10, $C = 0.5$, $\sigma = 1.2$, $\rho = 5.5$. Compared to the other methods, the restored image in figure 4-h shows a good quality improvement with a significantly enhanced edges and a good suppression of noise. The diffusion adequately follows the direction of curvature lines.

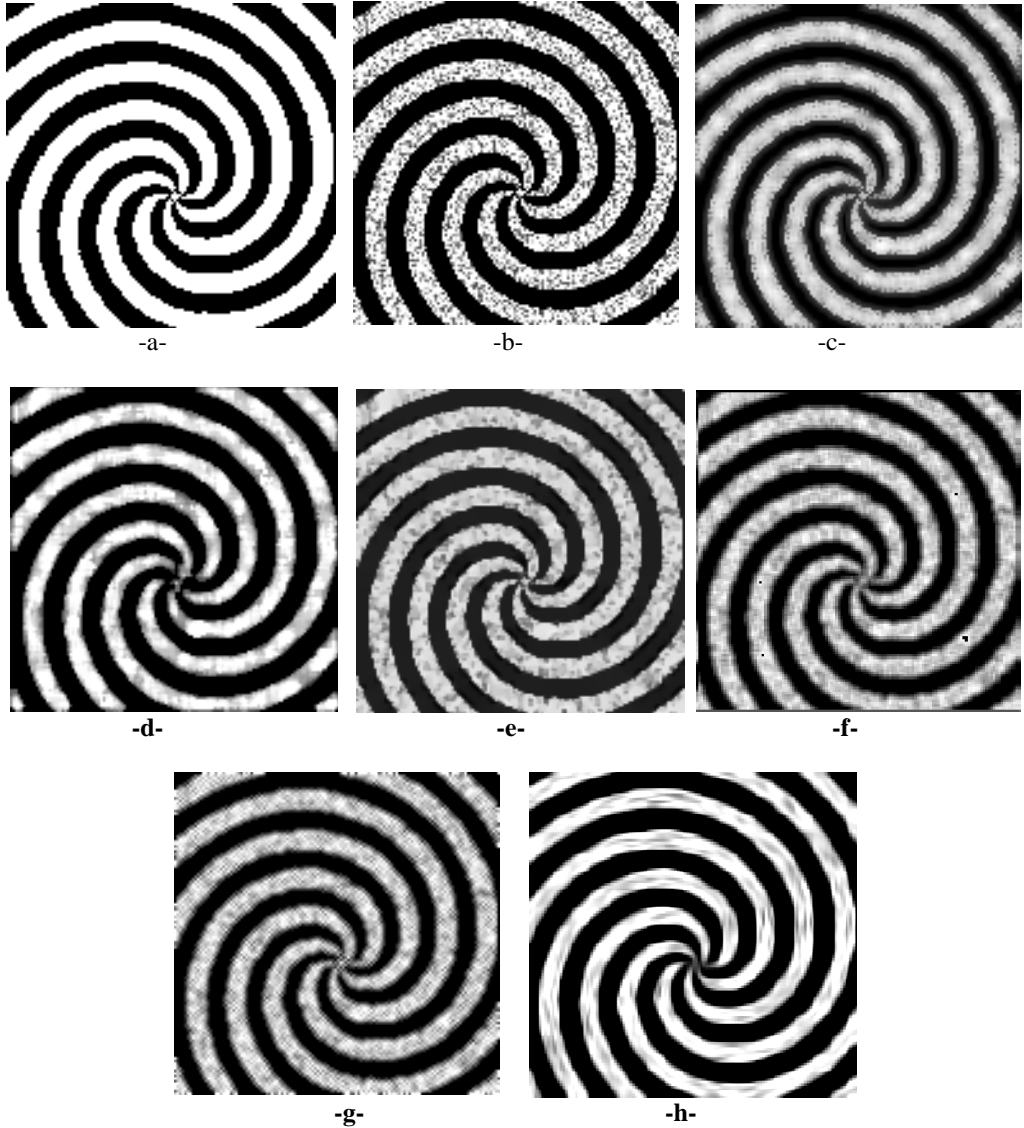

**Fig. 4** Image denoising simulation,-a- Original image,-b- Degraded image,-c- Lee method,-d- Median filter, -e-Total Variation (TV) ,-f-Kaun, -g- Frost,-h- Proposed method

To assess the quality image, we use two measures: the classic peak signal-to-noise-ratio (PSNR) and the mean structural similarity (MSSIM) index [15] which compares the structure of two images after subtracting luminance, and normalizing variance. The PSNR is defined by:

$$PSNR = 10 log_{10} \frac{N_{max}}{MSE} \qquad (17)$$

where $N_{max}$ : the maximum fluctuation in the input image , $N_{max} = (2^n-1)$, $N_{max} = 255$, when the components of a pixel are encoded on 8 bits; MSE : denotes the mean square error, given by :

$$MSE = \frac{1}{MN}\sum_{i=1}^{N}\sum_{j=1}^{M}\left|f(i,j) - \hat{f}(i,j)\right|^2 \qquad (18)$$

Where f(i,j) : the original image, $\hat{f}(i,j)$ : the restored image.

**Table 1:** Comparative table

| Method | PSNR (dB) | MSE | MSSIM |
|---|---|---|---|
| Lee | 15.00 | 0.031 | 0.79 |
| Median | 13.62 | 0.043 | 0.71 |
| TV | 14.03 | 0.039 | 0.73 |
| Kaun | 13.53 | 0.044 | 0.78 |
| Frost | 13.83 | 0.041 | 0.74 |
| Proposed | 16.10 | 0.023 | 0.82 |

We note that the MSSIM approximates the perceived visual quality of an image better than PSNR. It takes values in [0,1] and increases as the quality increases.

Table 1 confirms the effectiveness of our model with the highest score of MSSIM. In the next experiment, we apply the proposed algorithm on a real cardiac ultrasound image from a medical database sized 150x100 pixels. The model's parameters are fixed to: Nb_iter=10 , $C = 0.8$ , $\sigma = 1.2$, $\rho = 5.5$ . This image appears poor quality, lower resolution and very noisy as shown in figure 5.a. It can be seen that the restored image in figure 5.c, shows a significantly improvement with a good delimited cardiac cavity. Edges and discontinuities have been well recovered and preserved.

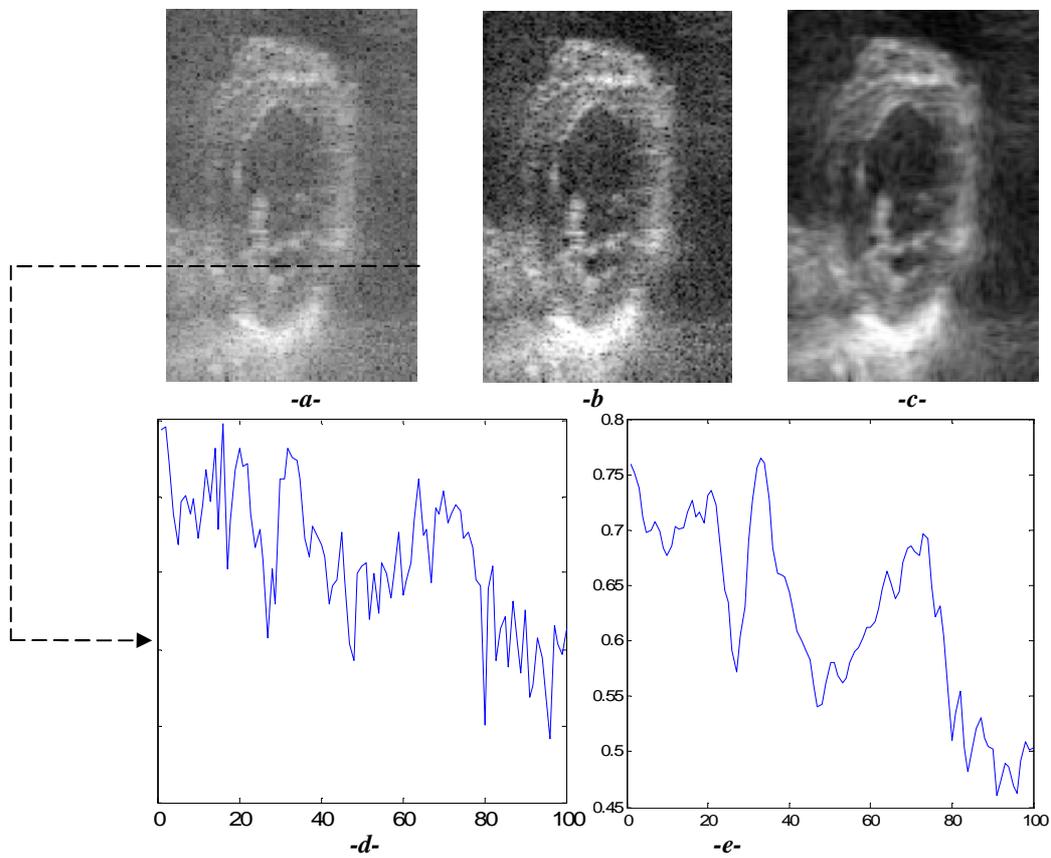

**Fig. 5** Results on ultrasound image, -a- Original image,-b- Contrast adjust of original image-c- Restored image, -d- Profile raw original image (raw=100),-e- Profile raw restored image.

## 6. Conclusion

The existence of speckle noise in the ultrasound image is undesirable since it disgrace image quality by affecting edges and local details between heterogeneous organs which are the most interesting part for diagnostics. In this paper, we have proposed a denoising approach which combines homomorphic transformation and diffusion tensors. The idea is to allow diffusion along the orientation of greatest coherence under condition of additive noise. However, the effectiveness of the proposed approach relies on the choice of the diffusion weight functions which control the diffusion along the orientation of greatest coherence. The experimental results on a real ultrasound images are very promising in terms of reducing speckle while preserving the appearance of structured regions and organ surfaces. This is very helpful to assist radiologist in their quest and diagnostic. Future works will include real-time speckle reduction and 3D ultrasound images denoising.

## Authors Biographies

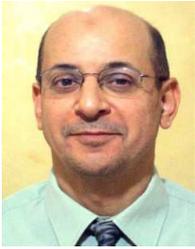

**Mr Faouzi Benzarti** received his Engineer's degree in Electrical Engineering from the Engineering School of Monastir (TUNISIA) in 1987, and his master's degree in Biomedical Engineering from the Polytechnic School of Montreal CANADA in 1991. He obtained his Ph.D degree from the Engineering School of Tunis (ENIT) in 2006. He is presently an assistant professor in the High School of Technology and Science of Tunis (ESSTT) and a member of research group in the Image Processing, Signal and Pattern Recognition TSRIF Laboratory. His current researches include: Image Deconvolution, Image Inpainting, Anisotropic diffusion, Image retrieval, 3D Biometry.

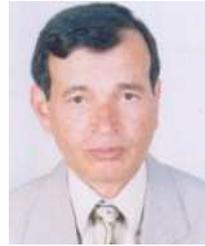

**Mr Hamid Amiri** received the Diploma of Electro-technics, Information Technique in 1978 and the PHD degree in 1983 at the TU Braunschweig, Germany. He obtained the Doctorates Sciences in 1993. He is presently a Professor at the National Engineering School of Tunis (ENIT) Tunisia. From 2001 to 2009 he was at the Riyadh College of Telecom and Information. Currently. He is now thea head member of research group in the Image, Signal and Pattern Recognition Laboratory. His research is focused on Image Processing, Speech Processing, Document Processing and Natural language processing.